\PassOptionsToPackage{table}{xcolor}
\documentclass[letterpaper]{article} % DO NOT CHANGE THIS
\usepackage{aaai2026}  % DO NOT CHANGE THIS
\usepackage{pdfpages} % 删
\usepackage{times}  % DO NOT CHANGE THIS
\usepackage{helvet}  % DO NOT CHANGE THIS
\usepackage{courier}  % DO NOT CHANGE THIS
\usepackage[hyphens]{url}  % DO NOT CHANGE THIS
\usepackage{graphicx} % DO NOT CHANGE THIS
\usepackage{natbib}  % DO NOT CHANGE THIS AND DO NOT ADD ANY OPTIONS TO IT
\usepackage{caption} % DO NOT CHANGE THIS AND DO NOT ADD ANY OPTIONS TO IT
\usepackage{algorithm}
\usepackage{amsfonts} % For \mathbb
\usepackage{amsmath}
\usepackage{algorithmic}

\usepackage{newfloat}
\usepackage{listings}
\DeclareCaptionStyle{ruled}
{labelfont=normalfont,labelsep=colon,strut=off} % DO NOT CHANGE THIS
\floatstyle{ruled}
\newfloat{listing}{tb}{lst}{}
\floatname{listing}{Listing}
\title{EgoPrune: Efficient Token Pruning for Egomotion Video Reasoning in Embodied Agent}
\author{
    Jiaao Li\textsuperscript{\rm 1},
    Kaiyuan Li\textsuperscript{\rm 1},
    Chen Gao\textsuperscript{\rm 2},
    Yong Li\textsuperscript{\rm 2},
    Xinlei Chen\textsuperscript{\rm 1}
}
\affiliations{
    \textsuperscript{\rm 1}Tsinghua Shenzhen International Graduate School\\
    \textsuperscript{\rm 2}Tsinghua University\\
    \{lja23, likaiyuan23\}@mails.tsinghua.edu.cn, \{chgao96, liyong07\}@tsinghua.edu.cn,  chen.xinlei@sz.tsinghua.edu.cn
}

\usepackage{bibentry}
\begin{document}

\maketitle

\begin{abstract}
Egomotion videos are first-person recordings where the view changes continuously due to the agent’s movement. As they serve as the primary visual input for embodied AI agents, making egomotion video reasoning more efficient is therefore essential for real-world deployment. Recent advances in vision-language models have enabled strong multimodal reasoning capabilities, but their computational cost remains prohibitive for long, redundant video inputs. Existing token pruning methods, typically designed for third-person videos, fail to leverage the spatiotemporal continuity and motion constraints inherent in egomotion settings. To address this, we propose EgoPrune, a training-free token pruning method tailored for egomotion video reasoning. EgoPrune comprises three components: a keyframe selector adapted from EmbodiedR for temporally efficient sampling; Perspective-Aware Redundancy Filtering (PARF), which aligns visual tokens using perspective transformations and removes redundant tokens; and a Maximal Marginal Relevance (MMR)-based token selector that jointly considers visual-text relevance and intra-frame diversity. Experiments on two egomotion video benchmarks show that EgoPrune consistently outperforms prior training-free methods across various pruning ratios while significantly reducing FLOPs, memory usage, and latency. Moreover, we deploy EgoPrune on an embodied agent equipped with a Jetson Orin NX 16GB edge device, demonstrating its real-world efficiency and suitability for on-device egomotion video reasoning.
\end{abstract}


\section{1 Introduction}
Recent advances in vision-language models (VLMs), such as LLaVA~\cite{liu2023visual} and Qwen-VL~\cite{bai2023qwen}, have enabled strong multimodal reasoning by integrating visual features into large language models. These models typically follow a two-stage architecture: a visual encoder extracts image or video frame features~\cite{radford2021learning}, which are fused with text tokens for joint processing. While effective, this approach introduces substantial computational costs, especially for videos, where multiple frames generate a large number of visual tokens. The resulting increase in token count burdens the model, as attention operations scale quadratically and KV cache usage grows linearly~\cite{vaswani2017attention,tay2022efficient,shi2024keep,jin2024efficient}.

Visual redundancy in videos exists at two levels: temporal (across frames) and spatial (within frames). Temporal redundancy arises due to frame-to-frame similarity~\cite{dutson2023eventful,chung2023shortcut}, while spatial redundancy results from repetitive patterns within individual frames~\cite{chen2024image,xing2024pyramiddrop}. Previous works address these issues by either removing temporally redundant tokens via cross-frame cosine similarity~\cite{yao2025timechat,li2025improving}, or selecting keyframes through uniform sampling or query-based retrieval~\cite{zhang2024video,tang2025adaptive}. For spatial redundancy, pruning methods rely on attention scores~\cite{lin2025boosting,meng2025plphp,yang2025visionzip} or intra-frame diversity to retain the most informative tokens~\cite{alvar2025divprune}.

Egomotion videos—captured from a first-person view where the camera moves continuously with the agent—are a central input modality in embodied AI~\cite{peirone2024backpack, lin2022egocentric}, enabling grounded spatial reasoning and decision-making~\cite{yang2025thinking, zhao2025urbanvideo}. Compared to third-person videos, egomotion videos exhibit strong temporal and spatial continuity due to physical constraints of agent motion, which introduce both opportunities and challenges for token pruning.

\begin{figure}[t]
    \centering
    \includegraphics[width=\columnwidth]{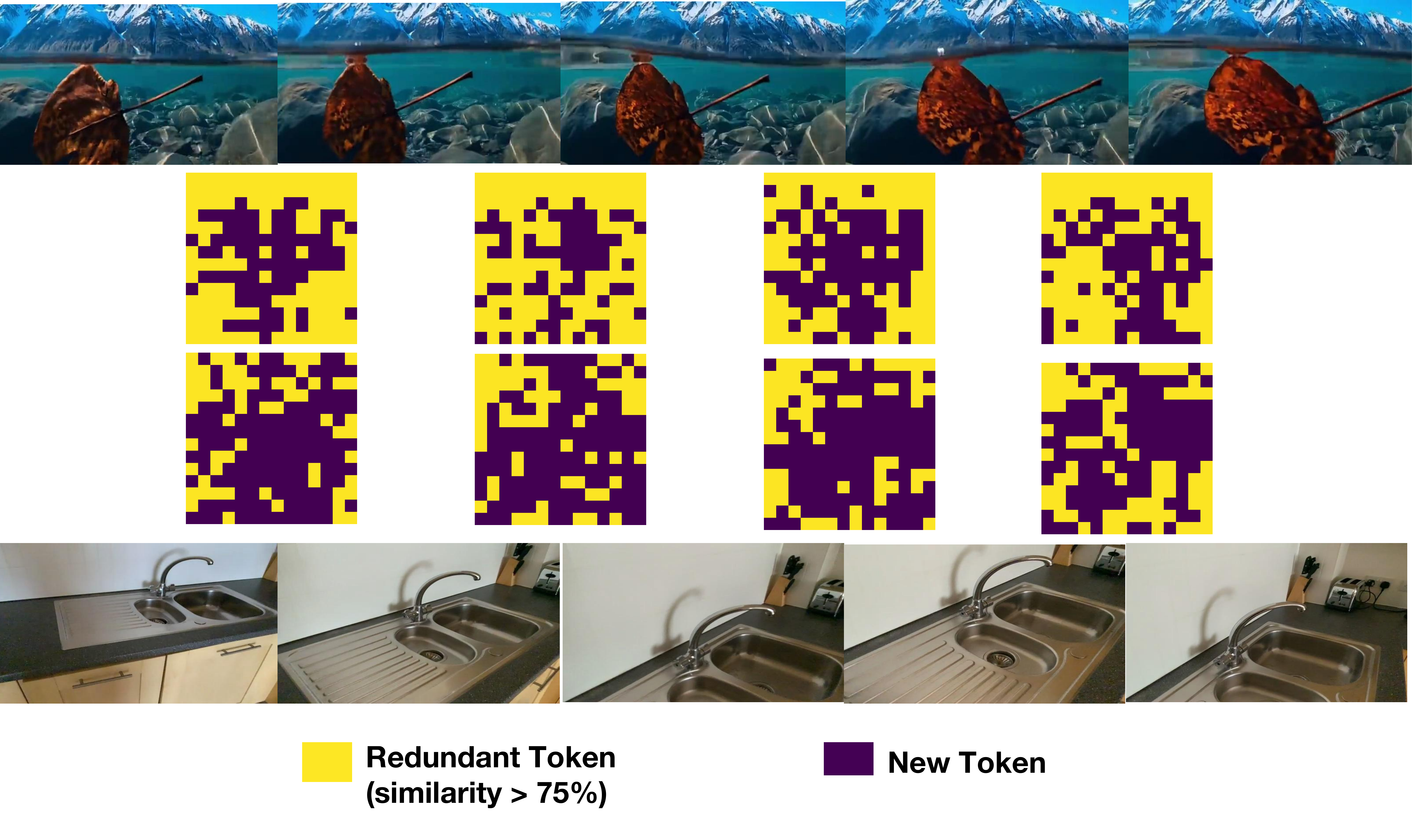}
    \caption{Identifying redundant tokens based on cosine similarity between corresponding positions in consecutive frames works for fixed cameras, but fails in egomotion videos where token alignment breaks due to viewpoint shifts.}
    \label{FIG:1}
\end{figure}

However, most existing pruning methods are designed for third-person perspectives and do not generalize well to egomotion settings. Fixed-position cosine similarity fails under continuous viewpoint shifts (see Figure~\ref{FIG:1}), and keyframe selection can discard critical transitional information. Attention-based methods are hindered by positional biases~\cite{wen2025token, li2025balanced}, incompatibility with optimized attention implementations like FlashAttention~\cite{dao2022flashattention}, and poor performance on perception-heavy tasks that demand spatial completeness. Diversity-based methods~\cite{alvar2025divprune} preserve representative visual information but are task-agnostic and may omit tokens essential for answering user queries~\cite{zhang2025beyond}.

Despite the significance of egomotion data, lightweight reasoning in this setting remains underexplored. Unlike third-person videos with frequent scene cuts, egomotion videos offer a smoother visual stream governed by motion continuity, making them ideal for structured, geometry-aware pruning.

\begin{figure*}[t]
    \centering
    \includegraphics[width=\textwidth]{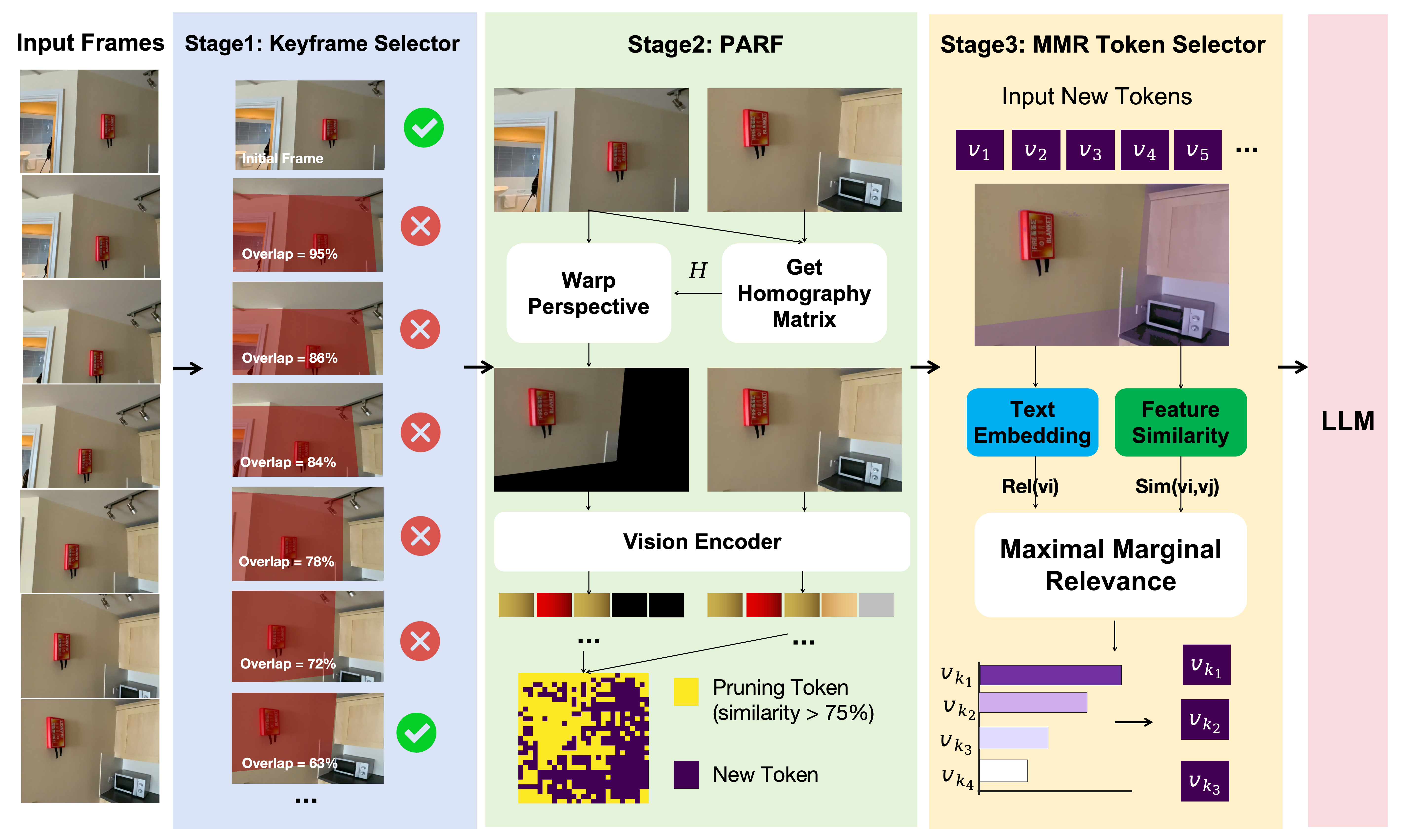}
    \caption{Overview of EgoPrune: Built on Embodied-R’s keyframe selection via overlap-based filtering, EgoPrune applies a two-stage token compression pipeline. Stage 1 (PARF) aligns consecutive frames using perspective transformation to prune redundant tokens. Stage 2 leverages Maximal Marginal Relevance (MMR) to select a subset of new tokens that are both relevant to the input text and diverse in visual semantics, improving inference efficiency while preserving task-relevant information.}
    \label{FIG:3}
\end{figure*}

To fill this gap, we introduce \textbf{EgoPrune}, a training-free token pruning method specifically designed for egomotion video reasoning. As shown in Figure~\ref{FIG:3}, EgoPrune consists of three components: (1) a keyframe selector adapted from EmbodiedR~\cite{zhao2025embodied} for temporally efficient sampling; (2) \textbf{Perspective-Aware Redundancy Filtering (PARF)}, which leverages homography transformations to align tokens across frames and filter redundant ones based on geometric correspondence; and (3) a \textbf{Maximal Marginal Relevance (MMR)} token selector that jointly considers prompt relevance and visual diversity, enabling informed and compact token retention~\cite{wu2023maximal}.

Extensive experiments on egomotion video benchmarks demonstrate that EgoPrune achieves competitive or superior performance compared to state-of-the-art methods, preserving over 99\% of task accuracy while significantly reducing FLOPs, memory usage, and latency. Furthermore, EgoPrune is successfully deployed on a Jetson Orin NX 16GB, validating its real-world efficiency for on-device embodied applications such as UAV navigation or mobile robotics.

\section{2 Related Work}
\subsection{2.1 Large Vision-Language Models (LVLMs)}
Large language models (LLMs) like GPT~\cite{achiam2023gpt} excel in text tasks but lack visual understanding due to their unimodal nature. Vision-language pretraining frameworks such as ViT~\cite{dosovitskiy2020image}, CLIP~\cite{radford2021learning}, and SigLIP~\cite{zhai2023sigmoid} bridge this gap by aligning images and text in a shared semantic space. Building on these, models like LLaVA~\cite{liu2023visual} and Qwen-VL~\cite{bai2023qwen} integrate visual encoders into LLMs via lightweight fusion layers to support vision-conditioned reasoning.

Recent advances extend vision-language modeling to video. Video-LLaMA~\cite{zhang2023video} adds a temporal encoder to LLaMA, while Video-ChatGPT~\cite{maaz2023video} adopts a video-adapted encoder with large-scale video instruction tuning. LLaVA-OneVision~\cite{li2024llava} unifies image and video processing via interleaved visual tokens and memory reuse. VILA~\cite{lin2024vila}, pretrained on interleaved image-text data, supports effective video reasoning and efficient edge deployment on devices like Jetson Orin.

\subsection{2.2 Training-free Video Token Pruning}
Video Large Language Models (VLLMs) show strong potential for video understanding but suffer from high inference cost due to the large number of visual tokens in long videos~\cite{jin2024efficient}, stemming from both temporal and spatial redundancy~\cite{dutson2023eventful,chung2023shortcut,chen2024image,xing2024pyramiddrop}. This motivates efficient token pruning without fine-tuning.

Recent methods address this by reducing token counts while preserving accuracy. TimeChat-Online~\cite{yao2025timechat} drops inter-frame redundant tokens, cutting 82.8\% with minimal loss. VLA-Cache~\cite{xu2025vla} reuses static tokens for up to 2.1× speed-up. DyCoke~\cite{tao2025dycoke} merges temporal and prunes spatial tokens for 1.7× acceleration. PACT~\cite{dhouib2025pact} combines early clustering and importance-based pruning (1.5× speed-up). DivPrune~\cite{alvar2025divprune} models token selection as a Max-Min Diversity Problem, achieving 52\% reduction while preserving structure.

\subsection{2.3 Keyframe Selector for Egomotion Video}
Most existing approaches for egomotion keyframe selection adopt uniform sampling to select input frames \cite{suglia2024alanavlm}. To move beyond this naïve strategy,  EmbodiedR \cite{zhao2025embodied} proposes a geometric keyframe selection method based on perspective transformation. It estimates the visual overlap between adjacent frames using feature matching and robust alignment techniques, and selects a new keyframe when the overlap drops below a predefined threshold.

\section{3 Preliminary}

\subsection{3.1 Video Token Generation and Processing}

Given a video of $T$ RGB frames $F = \{f_t\}_{t=1}^T$, with $f_t \in \mathbb{R}^{H \times W \times 3}$, each frame is split into $N$ non-overlapping patches via a patching function $f_{\text{patch}}$:
\[
p_t = f_{\text{patch}}(f_t) = \{p_t^i\}_{i=1}^N, \quad p_t^i \in \mathbb{R}^{h \times w \times 3}
\]

The patches are encoded into $N$ visual tokens using an encoder $f_{\text{enc}}$:
\[
v_t = f_{\text{enc}}(p_t) \in \mathbb{R}^{N \times d}
\]

All frame-level tokens are concatenated to form the video token sequence:
\[
V = [v_1; v_2; \ldots; v_T] \in \mathbb{R}^{NT \times d}
\]

Let $S \in \mathbb{R}^{N_S \times d}$ and $Q \in \mathbb{R}^{N_Q \times d}$ denote the system and user prompts. The full input to the language model is:
\[
X = [S; V; Q] \in \mathbb{R}^{(N_S + NT + N_Q) \times d}
\]

The language model then generates a response sequence of $N_O$ tokens:
\[
\text{Response} = \text{LLM}(X) \in \mathbb{R}^{N_O \times d}
\]

\subsection{3.2 Perspective Transformation}

Perspective transformation maps a point $\mathbf{x} = [x, y, 1]^\top$ in one image to $\mathbf{x}' = [x', y', 1]^\top$ in another via a $3 \times 3$ homography matrix $H$:
\[
\mathbf{x}' \sim H\mathbf{x}
\]
where $\sim$ denotes equality up to scale in homogeneous coordinates. $H$ is defined as:
\[
H =
\left[
\begin{array}{ccc}
h_1 & h_2 & h_3 \\
h_4 & h_5 & h_6 \\
h_7 & h_8 & h_9
\end{array}
\right]
\]

The homography assumption holds when the scene is planar or camera motion is purely rotational. To estimate $H$, at least four point correspondences $\mathbf{x}_i \leftrightarrow \mathbf{x}'_i$ are needed. Each correspondence yields two constraints:
\[
x'_i(h_7 x_i + h_8 y_i + h_9) = h_1 x_i + h_2 y_i + h_3
\]
\[
y'_i(h_7 x_i + h_8 y_i + h_9) = h_4 x_i + h_5 y_i + h_6
\]

Stacking all constraints yields a linear system $Ah = 0$, where $h = [h_1, \ldots, h_9]^\top$. The solution is obtained via SVD. In practice, more than four correspondences are used, with RANSAC~\cite{cantzler1981random} to reject outliers. Homography is widely used in stitching, motion estimation, and aligning egocentric frames where geometric consistency matters.

\subsection{3.3 Maximal Marginal Relevance}

Maximal Marginal Relevance (MMR)~\cite{wu2023maximal} is a greedy algorithm that selects a subset of $k$ items from $n$ candidates by balancing task relevance and intra-set diversity. It starts with the most relevant item, then iteratively adds the item that maximizes a trade-off between its relevance and similarity to already selected items.

Given relevance score $\text{reward}_i$ and pairwise similarity $\text{sim}(i, j)$, the marginal score for candidate $i$ is:
\[
\text{MR}_i = \lambda \cdot \text{reward}_i - (1 - \lambda) \cdot \max_{j \in \mathcal{S}} \text{sim}(i, j)
\]
where $\mathcal{S}$ is the current selection set and $\lambda \in [0, 1]$ controls the relevance-diversity balance.

The full procedure is summarized in Algorithm \ref{alg:mmr}.

\begin{algorithm}[tb]
\caption{Maximal Marginal Relevance (MMR)}
\label{alg:mmr}
\begin{algorithmic}[1]
\STATE \textbf{Input:} Relevance scores $\{\text{reward}_i\}_{i=1}^n$; similarity function $\text{sim}(i, j)$; selection count $k$; trade-off parameter $\lambda \in [0, 1]$
\STATE \textbf{Output:} Selected index set $\mathcal{S}$
\STATE Initialize selected set $\mathcal{S} \leftarrow \emptyset$, remaining set $\mathcal{R} \leftarrow \{1, 2, \ldots, n\}$
\STATE Select most relevant item: $i^\star \leftarrow \arg \max_{i \in \mathcal{R}} \text{reward}_i$
\STATE $\mathcal{S} \leftarrow \mathcal{S} \cup \{i^\star\}$, $\mathcal{R} \leftarrow \mathcal{R} \setminus \{i^\star\}$
\FOR{$t = 1$ to $k - 1$}
    \FOR{each $i \in \mathcal{R}$}
        \STATE Compute marginal relevance score: $\text{MR}_i = \lambda \cdot \text{reward}_i - (1 - \lambda) \cdot \max_{j \in \mathcal{S}} \text{sim}(i, j)$
    \ENDFOR
    \STATE $i^\star \leftarrow \arg\max_{i \in \mathcal{R}} \text{MR}_i$
    \STATE $\mathcal{S} \leftarrow \mathcal{S} \cup \{i^\star\}$, $\mathcal{R} \leftarrow \mathcal{R} \setminus \{i^\star\}$
\ENDFOR
\STATE \textbf{return} $\mathcal{S}$
\end{algorithmic}
\end{algorithm}

\section{4 Methodology}

\subsection{4.1 Limitations of existing methods}
Existing pruning methods assume redundant tokens remain spatially stable across frames~\cite{yao2025timechat,xu2025vla,li2025improving,tao2025dycoke}, which fails in egomotion videos where viewpoint shifts move even background regions, causing erroneous pruning of spatially essential tokens. Egomotion also exhibits continuous, physically constrained motion patterns that current methods overlook.

Spatial pruning faces further challenges. Attention-based methods~\cite{lin2025boosting,chen2024image,yang2025visionzip} suffer from RoPE-induced positional bias, incompatibility with efficient inference (e.g., FlashAttention-2), and poor performance on perception-heavy tasks requiring spatial completeness~\cite{wen2025token}. Diversity-based methods like DivPrune~\cite{alvar2025divprune} encourage token diversity but ignore task relevance, resulting in query-agnostic pruning. These issues call for egomotion-specific methods that integrate geometric consistency, perception relevance, and task adaptability.

\subsection{4.2 Our Method: EgoPrune}
% EgoPrune is built on EmbodiedR’s overlap-based keyframe selector \cite{zhao2025embodied} and adopts a two-stage token compression strategy. The first stage, \textbf{Perspective-Aware Redundancy Filtering (PARF)}, aligns consecutive keyframes via perspective transformation to prune redundant tokens with high similarity. The second stage, \textbf{MMR Token Selector}, applies Maximal Marginal Relevance to select a subset of newly visible tokens by jointly considering their relevance to the input text and their visual dissimilarity, ensuring both semantic importance and structural diversity.

\subsubsection{4.2.1 Perspective-Aware Redundancy Filtering (PARF)}
After applying Embodied-R’s overlap-aware keyframe selection, adjacent frames in egomotion videos typically retain 50\%–60\% visual overlap. This overlap captures smooth viewpoint transitions, helping the model incrementally construct a spatial understanding of the environment. However, tokens in overlapping regions are often highly redundant and impose unnecessary computational cost. Since video-language models use temporally-aware position encoding (e.g., 3D RoPE), the retained tokens across frames still carry distinct temporal embeddings, enabling the model to perceive motion without preserving all redundant tokens.

To address this, we align the previous frame to the current one via perspective transformation. Specifically, we estimate a homography matrix $H$ that maps previous frame to next frame based on geometric correspondences. ORB keypoints and descriptors are extracted, and matches are selected using FLANN and Lowe’s ratio test. RANSAC is then used to compute a robust estimate of $H$ from the matched keypoints.

\begin{figure}[t]
    \centering
    \includegraphics[width=\columnwidth]{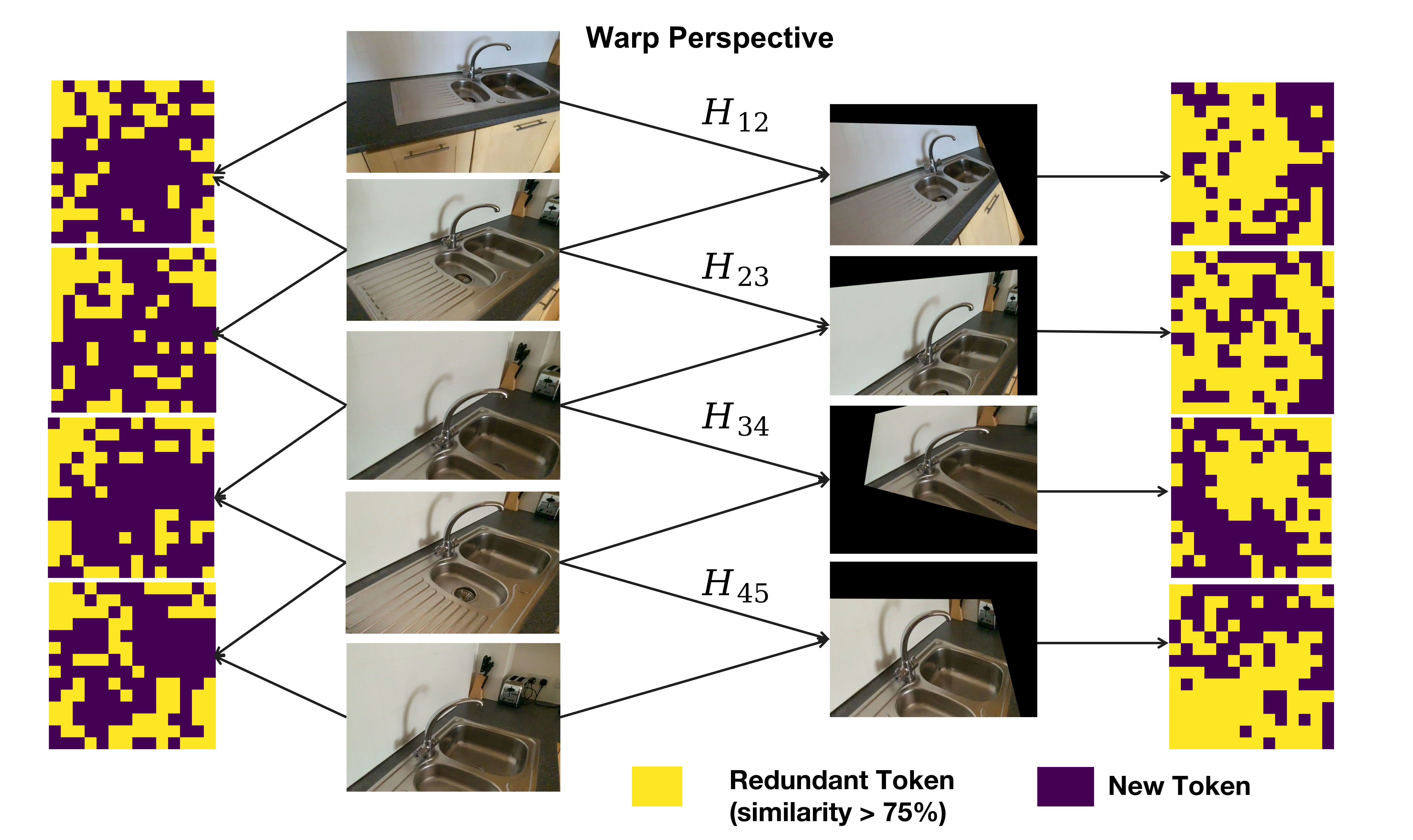}
    \caption{By estimating a homography matrix between consecutive frames, we warp the previous frame to align with the current one. This alignment enables spatial correspondence between tokens, making it easier to identify redundant tokens via cosine similarity.}
    \label{FIG:4}
\end{figure}

Once $H$ is obtained, we perform a perspective warp of the previous frame to align with the current one (Figure~\ref{FIG:4}). This alignment significantly increases local token redundancy, allowing us to compare aligned token pairs. Tokens with cosine similarity above 75\% are considered redundant and discarded.

The overall time complexity of PARF per keyframe pair is efficient: computing $H$ via ORB + RANSAC costs $O(K \log K)$ (with $K=500$–$1000$ keypoints), and similarity computation between $N$ token pairs of dimension $d$ costs $O(Nd)$. Thus, the total cost per pair is $O(K \log K + Nd)$, which is lightweight even when applied across dozens of frames.

% By leveraging homography-based alignment, PARF effectively filters redundant inter-frame tokens while preserving spatial continuity, offering a principled and low-overhead solution tailored for egomotion video.

\subsubsection{4.2.2 MMR Token Selector}
After obtaining the visual token sequence $V = [v_1; v_2; \ldots; v_T] \in \mathbb{R}^{NT \times d}$ and user prompt tokens $Q = [q_1; \ldots; q_{N_Q}] \in \mathbb{R}^{N_Q \times d}$, we apply Maximal Marginal Relevance (MMR) to select visual tokens that are both relevant to the prompt and diverse.

We define cosine similarity as:
$$
\text{sim}(v_i, v_j) = \frac{v_i^\top v_j}{|v_i| \cdot |v_j|}
$$

Following CDPruner \cite{zhang2025beyond}, the user prompt representation is computed as:
$$
q_{\text{avg}} = \frac{1}{N_Q} \sum_{j=1}^{N_Q} q_j
$$

The relevance of each visual token $v_i$ is:
$$
\text{rel}(v_i) = \text{sim}(v_i, q_{\text{avg}})
$$

MMR iteratively selects a subset $\mathcal{S} \subseteq {\{v_1, \ldots, v_{NT}}\}$ of size $k$ by maximizing:
$$
\text{MR}_i = \lambda \cdot \text{rel}(v_i) - (1 - \lambda) \cdot \max_{v_j \in \mathcal{S}} \text{sim}(v_i, v_j)
$$
where $\lambda \in [0, 1]$ balances relevance and diversity.

\begin{figure}[t]
\centering
\includegraphics[width=\linewidth]{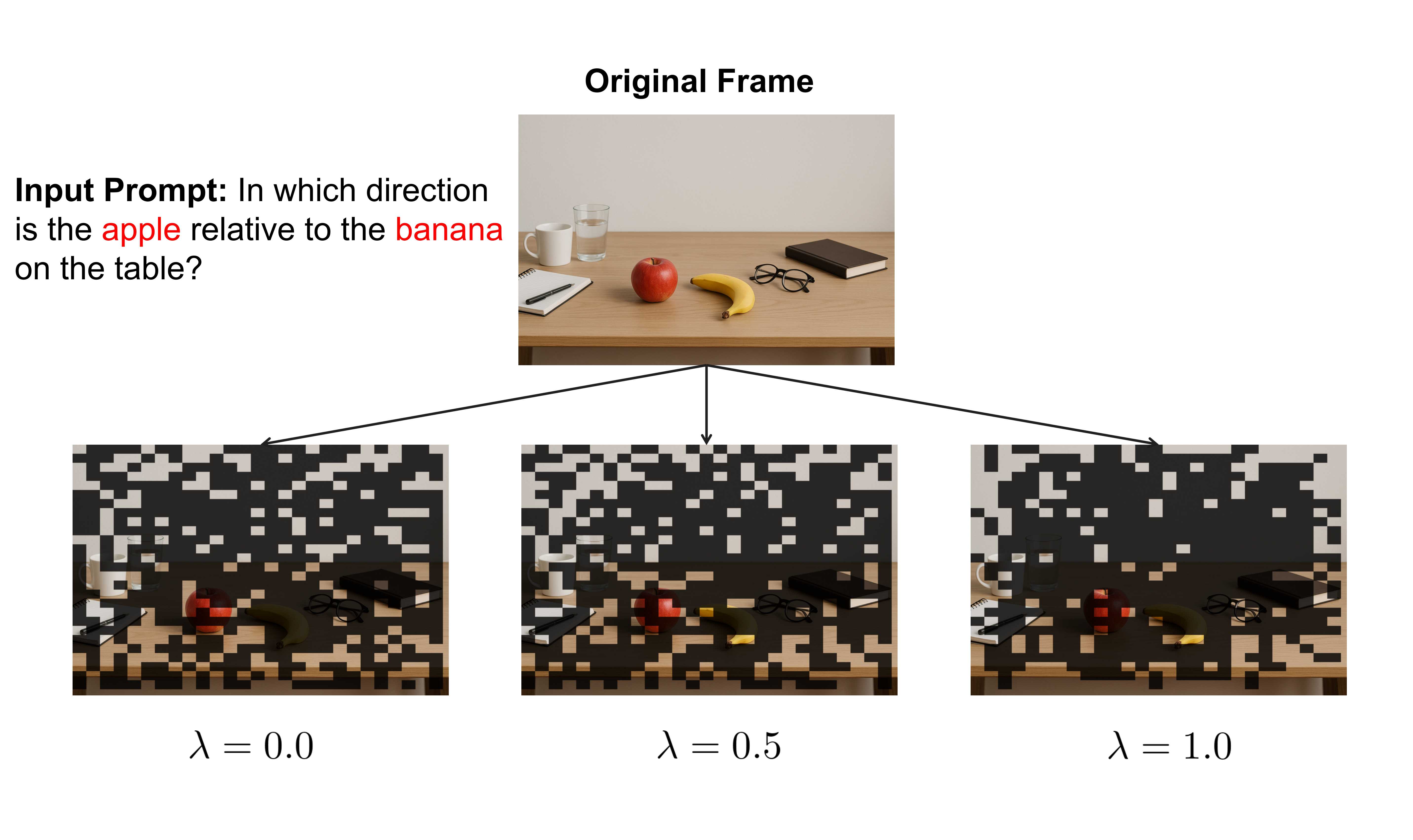}
\caption{Visualization of MMR Token Selector under different $\lambda$ settings. Larger $\lambda$ emphasizes relevance to the prompt, smaller $\lambda$ favors spatial diversity.}
\label{FIG:8}
\end{figure}

Figure~\ref{FIG:8} visualizes the effect of the balance factor $\lambda$. At $\lambda = 0$, tokens are spatially diverse but miss key semantics (e.g., no reserved tokens on banana). At $\lambda = 1$, the selection is highly prompt-relevant but lacks spatial coverage. $\lambda = 0.5$ achieves a better balance.

The computational complexity is $O(rwn^2)$, where $n = NT$ is the number of visual tokens, $r$ is the retained ratio, and $w = 10$ is the window size for computing diversity using only the most recent selections in $\mathcal{S}$. As $|\mathcal{S}|$ grows, recent tokens are sufficient to ensure diversity. With constant $r$ and $w$, the complexity simplifies to $O(n^2)$.

This approach ensures retained tokens are both semantically informative and spatially representative, without relying on attention scores—making it robust to positional bias and fully compatible with efficient attention implementations like FlashAttention-2.

\section{5 Experiments}
\subsection{5.1 Evaluation Setups and Implementation Details}

\subsubsection{5.1.1 Benchmarks}
We evaluate EgoPrune on two egomotion spatial reasoning benchmarks: \textbf{VSI-Bench}~\cite{yang2025thinking} and \textbf{UrbanVideo-Bench}~\cite{zhao2025urbanvideo}. VSI-Bench features 288 indoor egocentric videos and 5K+ QA pairs for embodied indoor reasoning. UrbanVideo-Bench comprises 1K aerial clips and 5.2K QA items, targeting city-scale spatial and directional understanding. Together, they cover diverse geometric and perceptual challenges across indoor and outdoor environments for embodied agents.

\subsubsection{5.1.2 Comparison Methods}
We compare EgoPrune with two training-free state-of-art token pruning methods: \textbf{DivPrune}~\cite{alvar2025divprune}, which solves a Max-Min Diversity Problem to select structurally diverse tokens, and \textbf{PACT}~\cite{dhouib2025pact}, which combines attention-free importance scoring with density-based clustering for FlashAttention-compatible pruning.

\subsubsection{5.1.3 Implementation Details}
We implement EgoPrune on LLaVA-OneVision-7B using PyTorch with three NVIDIA A6000 GPUs (50GB). All methods retain 70\%, 50\%, and 30\% of visual tokens for decoding. Videos are sampled at 1 FPS, with each frame yielding 196 tokens. To reflect real-world constraints, we filter out videos longer than 90 seconds. This is based on the assumption that embodied agents need to respond quickly to environmental changes, and the number of frames available between consecutive actions is typically limited. Evaluations are conducted with LMMS-Eval~\cite{zhang2024lmms}.

\subsection{5.2 Main Results}

\subsubsection{5.2.1 Effectiveness Evaluation}
As shown in Table~\ref{tab:token_retention_vsi} and Table~\ref{tab:token_retention_urban}, EgoPrune consistently outperforms existing training-free pruning methods across both VSI-Bench and UrbanVideo-Bench, demonstrating strong generalization from indoor embodied environments to large-scale urban aerial scenarios.

On VSI-Bench, EgoPrune achieves the highest average accuracy across all retention rates. At 70\% retention, it closely approaches the full-token baseline (35.20 vs. 35.45), and at 50\% and 30\%, it even exceeds the baseline (35.43 and 35.16), showing strong resilience to pruning. It also achieves top scores in key perception-driven tasks such as Object Appearance (27.9 at 30\%), Size Estimation (50.6), and Counting (up to 55.5), demonstrating its ability to retain spatially precise and object-relevant information under compression.

On UrbanVideo-Bench, EgoPrune matches or even surpasses the performance of the full-token baseline. At 70\% retention, it matches the baseline accuracy (46.25) while reducing computation, and maintains the highest accuracy under 50\% (46.16) and 30\% (45.66) pruning, consistently outperforming DivPrune and PACT. It also excels in high-level semantic tasks like Goal Detection (27.8), Progress Estimation (32.8), and Trajectory Captioning (39.9 at 30\%), highlighting its strength in preserving global scene semantics and temporal coherence for embodied reasoning.

\begin{table*}[t]
\centering
\caption{\textbf{Comparison of different methods on VSI-Bench.} For all the values, the higher is better. This table reports the accuracy across various spatial reasoning tasks, comparing multiple training-free pruning methods under LLaVA-OneVision-7B. The best-performing method for each task is highlighted in bold. \textit{Task definitions}: \textbf{Obj Appear} = Object Appearance Order; \textbf{Abs Dist} = Absolute Distance Estimation; \textbf{Count} = Object Counting; \textbf{Rel Dist} = Relative Distance Estimation; \textbf{Size Est} = Object Size Estimation; \textbf{Room Est} = Room Size Estimation; \textbf{Route} = Route Planning; \textbf{Rel Dir} = Relative Direction Understanding.}
\label{tab:token_retention_vsi}
\renewcommand{\arraystretch}{1.2}
% \resizebox{\textwidth}{!}{
\begin{tabular}{l|c|c|c|c|c|c|c|c|c}
\hline
\textbf{Methods} & \textbf{Avg.} & \textbf{Obj Appear} & \textbf{Abs Dist} & \textbf{Count} & \textbf{Rel Dist} & \textbf{Size Est} & \textbf{Room Est} & \textbf{Route} & \textbf{Rel Dir} \\
\hline
\rowcolor{gray!30}
Full Tokens       & 35.45 & 24.6 & 23.3 & 56.9 & 47.2 & 48.8 & 16.7 & 32.7 & 33.5  \\
\rowcolor{blue!10}
\multicolumn{10}{c}{\textbf{\textit{Retaining 70\% Tokens}}} \\
DivPrune (CVPR2025)  & 34.82 & 24.6 & \textbf{22.5} & 54.3 & 46.0 & 49.8 & 15.0 & 32.7 & \textbf{33.7}\\
PACT (CVPR2025) & 33.36 & 24.6 & 21.0 & 53.3 & 41.7 & 48.8 & \textbf{18.4} & 30.7 & 28.4 \\
\rowcolor{green!15}
\textbf{EgoPrune ($\lambda=0.5$)}  & \textbf{35.20} & \textbf{26.2} & 20.8 & \textbf{55.5} & \textbf{46.0} & \textbf{50.6} & 17.6 & \textbf{32.7} & 32.9 \\

\rowcolor{blue!10}
\multicolumn{10}{c}{\textbf{\textit{Retaining 50\% Tokens}}} \\
DivPrune (CVPR2025)   & 34.15 & 23.0 & \textbf{24.1} & 54.4 & 44.2 & \textbf{49.0} & 14.9 & 30.7 & \textbf{33.0} \\
PACT (CVPR2025) & 33.67 & 26.2 & 20.3 & 51.0 & 39.9 & 48.9 & \textbf{19.6} & 32.7 & 30.8 \\
\rowcolor{green!15}
\textbf{EgoPrune ($\lambda=0.5$)}  & \textbf{35.43} & \textbf{26.2} & 21.9 & \textbf{54.6} & \textbf{48.5} & 48.8 & 18.1 & \textbf{32.7} & 32.6 \\

\rowcolor{blue!10}
\multicolumn{10}{c}{\textbf{\textit{Retaining 30\% Tokens}}} \\
DivPrune (CVPR2025)   & 34.50 & 24.6 & \textbf{23.2} & \textbf{55.7} & 44.8 & 49.0 & 17.5 & 29.7 & 31.6 \\
PACT (CVPR2025) & 32.09 & 24.6 & 22.6 & 45.4 & 41.1 & 48.8 & 15.2 & 29.7 & 29.3 \\
\rowcolor{green!15}
\textbf{EgoPrune ($\lambda=0.5$)}  & \textbf{35.16} & \textbf{27.9} & 22.3 & 48.7 & \textbf{46.6} & \textbf{50.0} & \textbf{23.2} & \textbf{30.7} & \textbf{31.8} \\
\hline
\end{tabular}
% }
\end{table*}

\begin{table*}[t]
\centering
\caption{\textbf{ Comparison of different methods on UrbanVideo-Bench.} For all the values, the higher is better. The best-performing method for each task is highlighted in bold. This table reports results on a subset of spatial reasoning tasks; for the complete set of tasks, please refer to the Appendix. \textit{Task definitions}: \textbf{ActGen} = Action Generation; \textbf{CogMap} = Cognitive Map Construction; \textbf{CF} = Counterfactual; \textbf{GDet} = Goal Detection; \textbf{Progress} = Progress Evaluation; \textbf{Proximity} = Proximity Estimation; \textbf{SeqRecall} = Sequence Recall; \textbf{TrajCap} = Trajectory Caption.}
\label{tab:token_retention_urban}
\renewcommand{\arraystretch}{1.2}
% \resizebox{\textwidth}{!}{
\begin{tabular}{l|c|c|c|c|c|c|c|c|c}
\hline
\textbf{Methods} & \textbf{Avg.} & \textbf{ActGen} & \textbf{CogMap} & \textbf{CF} & \textbf{GDet} & \textbf{Progress} & \textbf{Proximity} & \textbf{SeqRecall} & \textbf{TrajCap} \\
\hline
\rowcolor{gray!30}
Full Tokens & 46.23 & 16.0 & 51.4 & 43.4 & 27.4 & 30.3 & 56.1 & 49.2 & 35.6  \\
\rowcolor{blue!10}
\multicolumn{10}{c}{\textbf{\textit{Retaining 70\% Tokens}}} \\
DivPrune (CVPR2025)   & 45.62 & \textbf{15.4} & \textbf{52.4} & 39.5 & \textbf{27.4} & 29.7 & 62.1 & \textbf{50.8} & 36.9 \\
PACT (CVPR2025) & 44.97 & 14.9 & 50.5 & 40.8 & 26.6 & \textbf{32.3} & 56.1 & 43.1 & \textbf{39.9} \\
\rowcolor{green!15}
\textbf{EgoPrune ($\lambda=0.5$)}  & \textbf{46.25} & 14.9 & 51.9 & \textbf{40.8} & 25.4 & 31.3 & \textbf{62.1} & 49.2 & 39.9 \\

\rowcolor{blue!10}
\multicolumn{10}{c}{\textbf{\textit{Retaining 50\% Tokens}}} \\
DivPrune (CVPR2025)   & 45.37 & \textbf{15.2} & 51.9 & 42.1 & 25.8 & 30.8 & 57.6 & 46.2 & 38.9 \\
PACT (CVPR2025) & 45.55 & 15.0 & 51.0 & 42.1 & \textbf{28.2} & 31.3 & \textbf{59.1} & 46.2 & 39.5 \\
\rowcolor{green!15}
\textbf{EgoPrune ($\lambda=0.5$)}  & \textbf{46.16} & 14.1 & \textbf{53.3} & \textbf{44.7} & 27.4 & \textbf{31.8} & 57.6 & \textbf{46.2} & \textbf{39.9} \\
\rowcolor{blue!10}
\multicolumn{10}{c}{\textbf{\textit{Retaining 30\% Tokens}}} \\
DivPrune (CVPR2025)   & 44.98 & \textbf{15.8} & 50.0 & 44.7 & 23.0 & 31.8 & 56.1 & \textbf{53.8} & 37.9 \\
PACT (CVPR2025) & 45.13 & 15.2 & \textbf{51.9} & 40.8 & \textbf{27.8} & 30.8 & 54.5 & 49.2 & 38.9 \\
\rowcolor{green!15}
\textbf{EgoPrune ($\lambda=0.5$)}  & \textbf{45.66} & 14.3 & 49.5 & \textbf{44.7} & 25.8 & \textbf{32.8} & \textbf{59.1} & 49.2 & \textbf{38.9} \\
\hline
\end{tabular}
% }
\end{table*}

\subsubsection{5.2.2 Efficiency Evaluation}
\begin{figure}[t]
    \centering
    \includegraphics[width=\linewidth]{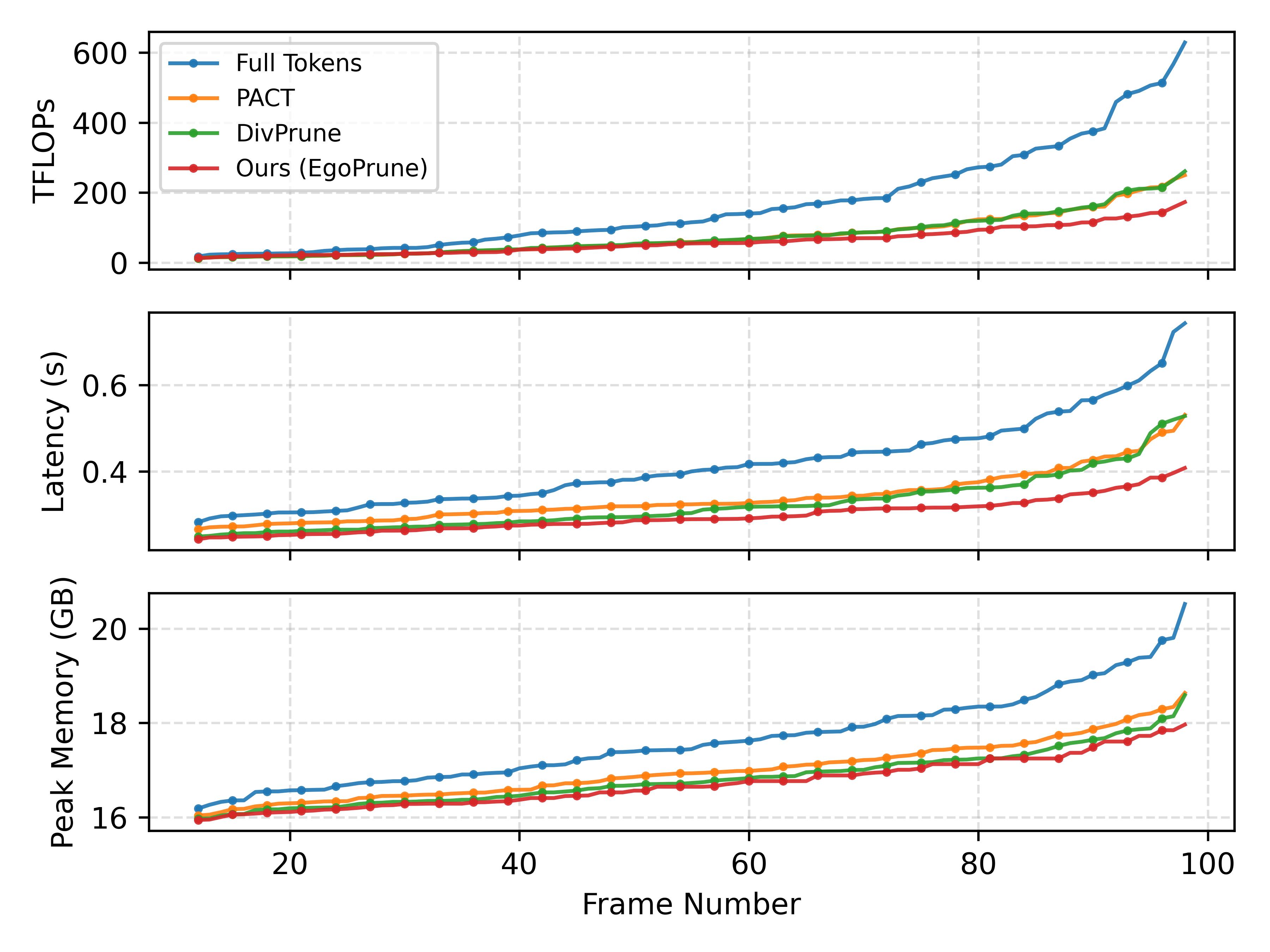}
    \caption{This figure shows the comparison of PACT, DivPrune, and EgoPrune under different frame counts in terms of TFLOPs (top), end-to-end latency (middle), and peak memory usage (bottom).}
    \label{FIG:5}
\end{figure}

We evaluate efficiency by measuring TFLOPs, end-to-end latency (averaged over 10 warm-up runs), and peak memory usage under varying input lengths, with all methods retaining 50\% of tokens for fair comparison. Metrics are collected using DeepSpeed Profiler~\cite{aminabadi2022deepspeed}, with all methods using Scaled Dot-Product Attention (SDPA) for consistency. Results are shown in Figure~\ref{FIG:5}.

EgoPrune consistently outperforms baselines across all metrics, especially on longer inputs. It incurs lower computational cost and memory usage, and exhibits smoother scaling in latency and TFLOPs—indicating better efficiency and token-wise scalability.

\subsection{5.3 Ablation Study}

\subsubsection{5.3.1 Ablation on Balance Factor $\lambda$}
\begin{figure}[t]
    \centering
    \includegraphics[width=\linewidth]{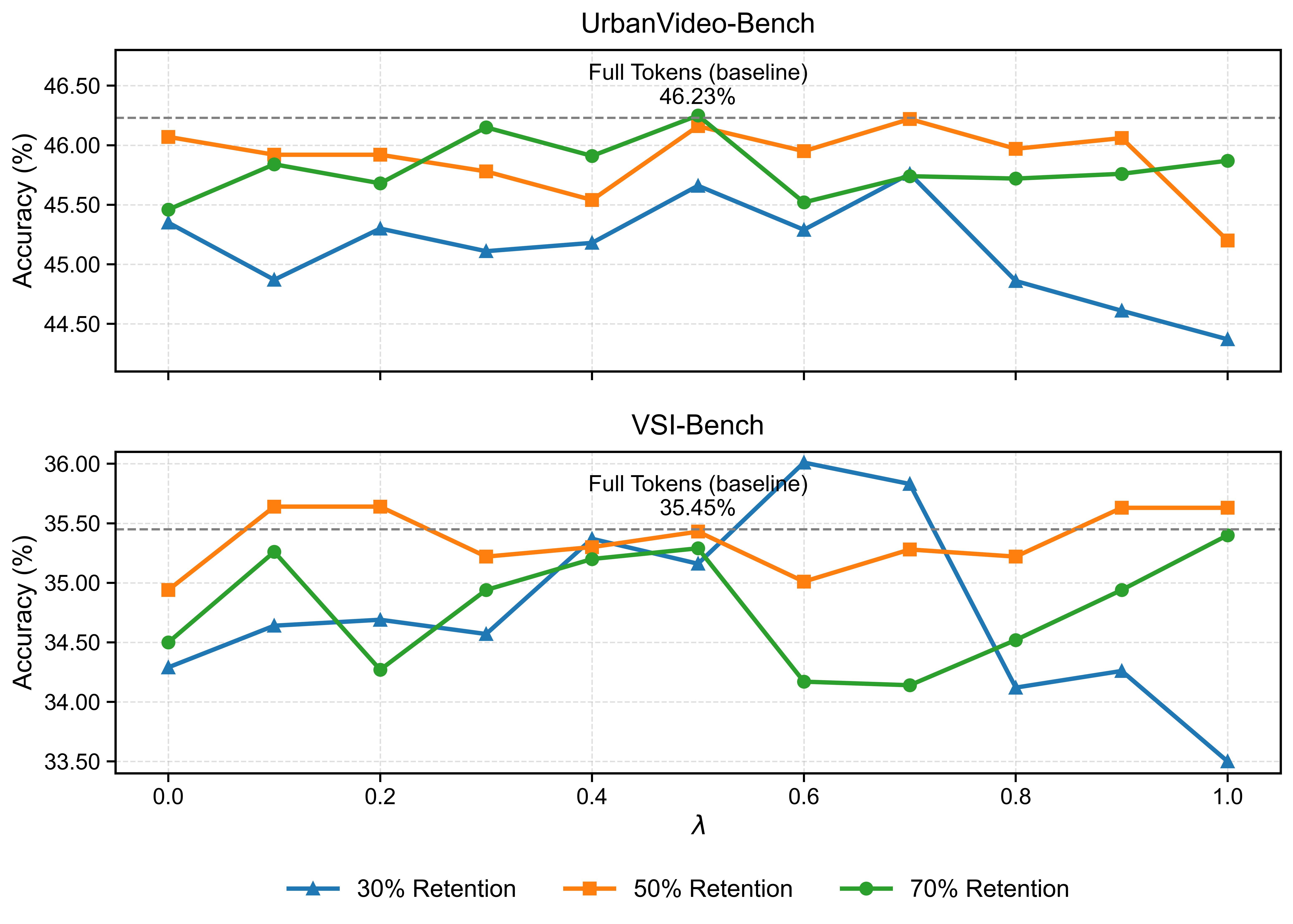}
    \caption{Performance of EgoPrune across varying values of the balance factor $\lambda$ in the MMR module, under 30\%, 50\%, and 70\% token retention. Results are reported on both UrbanVideo-Bench and VSI-Bench.}
    \label{FIG:7}
\end{figure}
Figure~\ref{FIG:7} shows the performance of EgoPrune on both UrbanVideo-Bench and VSI-Bench under varying values of the MMR balance factor $\lambda$, which controls the trade-off between prompt relevance and visual diversity during token selection.

On VSI-Bench, EgoPrune remains robust to $\lambda$ variations at 50\% and 70\% retention, with accuracy peaking at both low and high extremes (e.g., $\lambda=0.1$ and $\lambda=1.0$), indicating that either relevance or diversity can suffice in spatially redundant indoor environments. However, at 30\% retention, performance becomes more sensitive—accuracy drops sharply when $\lambda$ approaches 1.0, showing that overemphasizing diversity may sacrifice task-relevant content under aggressive pruning.

On UrbanVideo-Bench, performance consistently peaks at mid-range $\lambda$ values (e.g., 0.4–0.6), especially under 50\% and 70\% retention. This suggests that dynamic, outdoor UAV scenarios benefit from a balanced selection strategy. Extreme values (e.g., $\lambda=0.0$ or $\lambda=1.0$) underperform, particularly at 30\%, where relying solely on one criterion risks omitting key visual context.

Overall, optimal $\lambda$ depends on both pruning strength and scene type. Mid-range settings ($\lambda = 0.5$–$0.7$) yield stable performance across benchmarks, while slightly higher values (e.g., $0.6$–$0.8$) are preferable in structured, low-retention cases. EgoPrune shows strong robustness, making it effective with minimal tuning.

% \textbf{Recommendation for deployment:}
% \begin{itemize}
%     \item Use $\lambda = 0.5$ as a robust default across diverse settings.
%     \item For token-constrained scenarios, prioritize relevance with $\lambda \in [0.6, 0.8]$.
%     \item In open-world or egocentric scenarios, maintain balance with $\lambda \in [0.4, 0.6]$.
% \end{itemize}

\subsection{5.4 On-Device Evaluation}
To evaluate real-world efficiency, we deploy VILA-1.5 3B with EgoPrune on a Jetson Orin NX 16GB—an edge platform representative of embodied agents. EgoPrune is integrated into the inference pipeline, with both vision and language components compiled into TensorRT-LLM engines using INT4 weight-only quantization.

\begin{figure}[t]
    \centering
    \includegraphics[width=\linewidth]{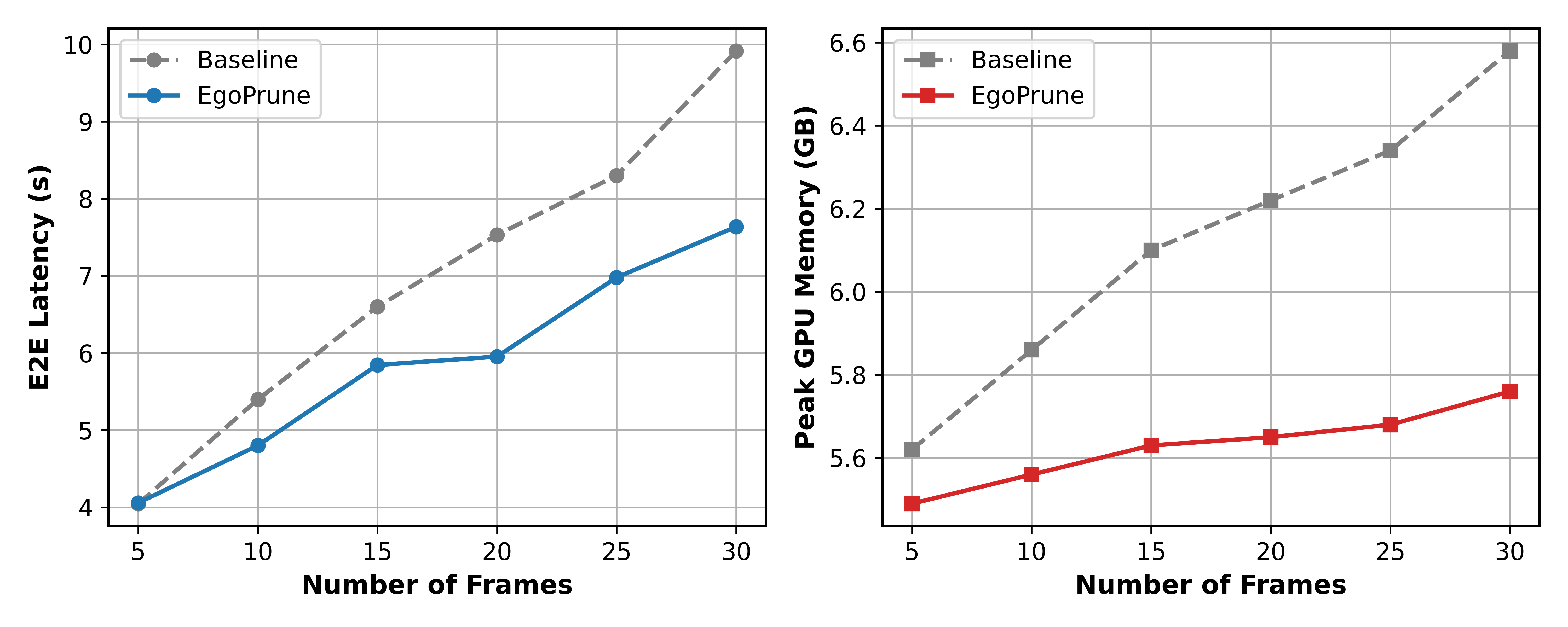}
    \caption{End-to-end latency (left) and peak GPU memory (right) on Jetson Orin NX 16GB for baseline and EgoPrune ($\lambda=0.5$, token retation rate is set to 50\%).}
    \label{FIG:11}
\end{figure}

As shown in Figure~\ref{FIG:11}, EgoPrune consistently reduces end-to-end latency across different frame counts, enabling faster processing within time-constrained scenarios. It also lowers peak GPU memory usage, supporting stable co-execution with onboard systems (e.g., UAV control) and mitigating Out-Of-Memory risks.

\begin{figure}[t]
    \centering
    \includegraphics[width=\linewidth]{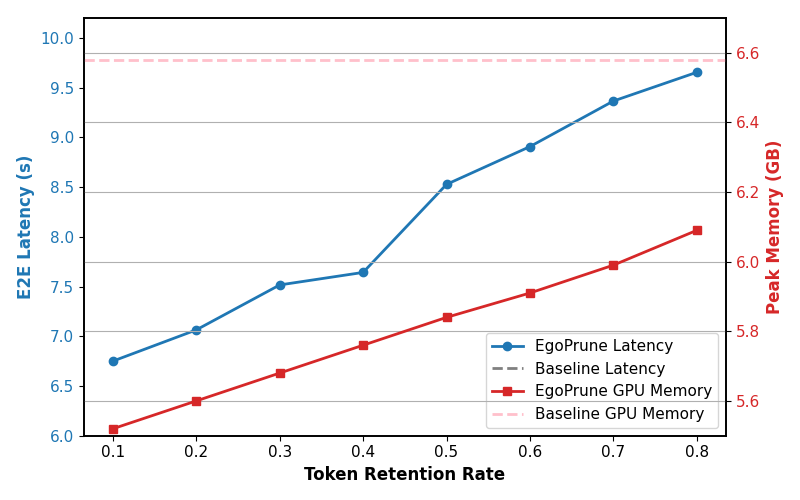}
    \caption{Token Retention Rate in MMR Token Selector ($\lambda = 0.5$) versus end-to-end latency and peak GPU memory.}
    \label{FIG:12}
\end{figure}

As shown in Figure~\ref{FIG:12}, we analyze how varying the token retention rate in the MMR Token Selector impacts latency and peak GPU memory. Memory usage scales roughly linearly with the retention rate. Interestingly, reducing the rate from 0.5 to 0.4 leads to a sharp latency drop~\cite{eliopoulos2025pruning}, indicating that tuning near such thresholds offers significant efficiency gains.

\section{6 Conclusion}
In this work, we present EgoPrune, a training-free token pruning framework designed for egomotion video reasoning in embodied agents. Built upon Embodied-R’s keyframe selection, EgoPrune introduces two key components: (1) Perspective-Aware Redundancy Filtering (PARF), which eliminates inter-frame redundancy via homography-based alignment, and (2) an MMR-based token selector, which jointly considers text relevance and visual diversity. Extensive experiments on both indoor (VSI-Bench) and outdoor (UrbanVideo-Bench) benchmarks demonstrate that EgoPrune consistently outperforms state-of-the-art training-free methods under various token reduction ratios, delivering competitive or superior accuracy while significantly improving efficiency in terms of FLOPs, end-to-end latency, and GPU memory usage. These results validate EgoPrune as an effective and lightweight solution for efficient egomotion video reasoning for embodied agent.

\bibliography{aaai2026}

\newpage



\section*{Supplementary material}

# Technical Appendix for EgoPrune: Efficient Token Pruning for Egomotion Video Reasoning in Embodied Agent

**Jiaao Li[1], Kaiyuan Li[1], Chen Gao[2], Yong Li[2], Xinlei Chen[1]**

[1]Tsinghua Shenzhen International Graduate School
[2]Tsinghua University
{lja23, likaiyuan23}@mails.tsinghua.edu.cn, {chgao96, liyong07}@tsinghua.edu.cn, chen.xinlei@sz.tsinghua.edu.cn

## Ablation on PARF and MMR Modules

Table 1 presents an ablation analysis of EgoPrune on both VSI-Bench and UrbanVideo-Bench. All results are reported under a fixed token retention rate of 50%.

On VSI-Bench, removing either PARF or MMR leads to noticeable performance drops, with average accuracy falling from 35.43 (full) to 34.82 (w/o PARF) and 34.55 (w/o MMR). The full model outperforms all variants in key spatial reasoning tasks like Object Appearance (26.2), Absolute Distance (21.9), and Route Estimation (32.7), showing that both modules are essential. Specifically, PARF is more effective in geometry-sensitive tasks (Abs Dist, Size Est), while MMR better preserves object-centric cues (Obj Appear, Rel Dir), highlighting their complementary roles.

On UrbanVideo-Bench, EgoPrune again achieves the highest average score (46.16), outperforming both ablations across most tasks. Trajectory Captioning, for example, drops to 37.9 (w/o PARF) and 38.6 (w/o MMR), confirming the importance of both components for long-range semantic consistency. While removing PARF slightly improves Proximity and SeqRecall, it harms cognitively demanding tasks like Goal Detection and CogMap, indicating a trade-off.

Figure 1 further supports these findings: removing PARF or MMR increases FLOPs, latency, and memory—especially with longer videos. PARF notably reduces FLOPs and memory by filtering redundant spatial tokens, while MMR lowers latency via relevance-aware selection. The full EgoPrune model consistently achieves the best efficiency, validating that the two modules offer complementary gains in both performance and resource savings.

## Additional Results on UrbanVideo-Bench

In the main text, we report results on a subset of UrbanVideo-Bench tasks for comparison and ablation studies. For completeness, the appendix presents full results across all tasks.

Table 2 shows ablation results on the remaining UrbanVideo-Bench tasks, focusing on visual reasoning such as Association Reasoning (AR), High-Level Planning (HL Plan), and Start/End Position (SEPos). The full model (MMR + PARF) achieves the best or competitive results across most tasks while retaining only 50% of tokens, notably outperforming variants on Causal (59.7), ObjRecall (70.7), and SEPos (64.7). Removing MMR reduces AR performance (16.9 →

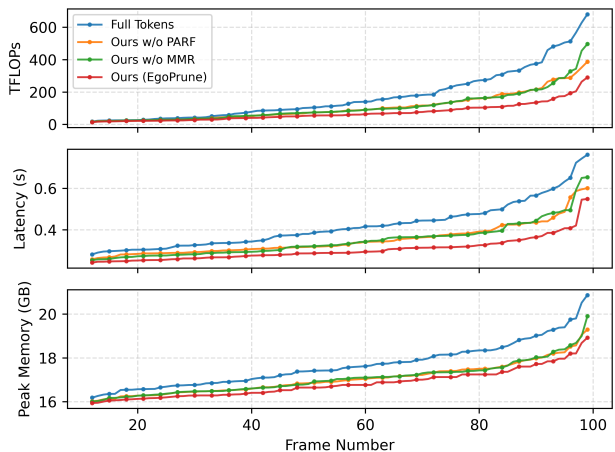

Figure 1: This figure illustrates the efficiency comparison among the full EgoPrune model and its two ablated variants (w/o PARF and w/o MMR) under varying input frame counts.

15.7), showing the importance of prompt-guided relevance, while removing PARF hurts HL Plan and SEPos, underscoring the role of token diversity. These results validate MMR's balance of relevance and diversity for challenging reasoning tasks.

Table 3 compares EgoPrune with DivPrune and PACT on the remaining UrbanVideo-Bench tasks under varying token retention ratios (70%, 50%, 30%). Tasks include high-level reasoning (Causal, Duration, HL Plan) and low-level perception (Landmark Pos, ObjRecall, SEPos). EgoPrune consistently achieves strong or best results, especially on ObjRecall and SEPos, highlighting its ability to preserve key spatial details. While DivPrune emphasizes visual diversity and PACT focuses on prompt relevance, both underperform on tasks requiring both aspects, such as Duration or Landmark Pos. In contrast, EgoPrune's two-stage design (PARF + MMR) balances relevance and spatial coverage, maintaining robust performance even under aggressive pruning.

Table 1: **Ablation Study of EgoPrune on VSI-Bench and UrbanVideo-Bench.** All results are reported at a 50% token retention rate for MMR Token Selector.

| Methods | Avg. | Obj Appear | Abs Dist | Count | Rel Dist | Size Est | Room Est | Route | Rel Dir |
|---|---|---|---|---|---|---|---|---|---|
| | | | | *VSI-Bench* | | | | | |
| Full Tokens | 35.45 | 24.6 | 23.3 | 56.9 | 47.2 | 48.8 | 16.7 | 32.7 | 33.5 |
| Ours w/o MMR | 34.55 | 23.0 | 19.9 | **55.8** | 45.4 | **50.1** | 18.0 | 31.7 | **32.6** |
| Ours w/o PARF ($\lambda = 0.5$) | 34.82 | 23.0 | 20.7 | 54.5 | **49.1** | 49.1 | **18.8** | 31.7 | 31.8 |
| **Ours** ($\lambda = 0.5$) | **35.43** | **26.2** | 21.9 | 54.6 | 48.5 | 48.8 | 18.1 | **32.7** | 32.6 |
| Methods | Avg. | ActGen | CogMap | CF | Goal Detect | Progress | Proximity | SeqRecall | TrajCaption |
| | | | | *UrbanVideo-Bench* | | | | | |
| Full Tokens | 46.23 | 16.0 | 51.4 | 43.4 | 27.4 | 30.3 | 56.1 | 49.2 | 35.6 |
| Ours w/o MMR | 45.91 | 15.4 | 51.4 | 40.8 | 25.8 | 30.8 | 59.1 | 49.2 | 38.6 |
| Ours w/o PARF ($\lambda = 0.5$) | 45.62 | **15.6** | 52.9 | 40.8 | 25.4 | 31.8 | **60.6** | **52.3** | 37.9 |
| **Ours** ($\lambda = 0.5$) | **46.16** | 14.1 | **53.3** | **44.7** | **27.4** | **31.8** | 57.6 | 46.2 | **39.9** |

Table 2: **Ablation Study of UrbanVideo-Bench on remaining tasks.** For all the values, the higher is better. The best-performing method for each task is highlighted in bold. All results are reported at a 50% token retention rate for MMR Token Selector. *Task definitions:* **AR** = Association Reasoning, **Causal** = Causal Inference, **HL Plan** = High-Level Planning, **Landmark Pos** = Landmark Positioning, **ObjRecall** = Object Recall, **SceneRecall** = Scene Recall, **SEPos** = Start/End Position.

| Methods | Avg. | AR | Causal | Duration | HL Plan | Landmark Pos | ObjRecall | SceneRecall | SEPos |
|---|---|---|---|---|---|---|---|---|---|
| | | | | *UrbanVideo-Bench* | | | | | |
| Full Tokens | 46.23 | 18.0 | 61.2 | 51.4 | 53.6 | 37.7 | 66.7 | 79.4 | 62.4 |
| Ours w/o MMR | 45.91 | 15.7 | 58.2 | 50.0 | 52.1 | 37.7 | 68.0 | **79.4** | 62.4 |
| Ours w/o PARF ($\lambda = 0.5$) | 45.62 | **18.0** | 58.2 | 45.8 | **53.1** | **38.5** | 64.0 | 76.2 | 58.8 |
| **Ours** ($\lambda = 0.5$) | **46.16** | 16.9 | **59.7** | **50.0** | 51.7 | 36.9 | **70.7** | 73.0 | **64.7** |

Table 3: **Comparison of different methods on UrbanVideo-Bench on remaining tasks.** For all the values, the higher is better. The best-performing method for each task is highlighted in bold.

| Methods | Avg. | AR | Causal | Duration | HL Plan | Landmark Pos | ObjRecall | SceneRecall | SEPos |
|---|---|---|---|---|---|---|---|---|---|
| Full Tokens | 46.23 | 18.0 | 61.2 | 51.4 | 53.6 | 37.7 | 66.7 | 79.4 | 62.4 |
| | | | | *Retaining 70% Tokens* | | | | | |
| DivPrune (CVPR2025) | 45.62 | 16.3 | 58.2 | 47.2 | **53.1** | 38.2 | **68.0** | 74.6 | 60.0 |
| PACT (CVPR2025) | 44.97 | 15.7 | **62.7** | 47.2 | 49.3 | **38.2** | 61.3 | 81.0 | 60.0 |
| **EgoPrune** ($\lambda = 0.5$) | **46.25** | **16.3** | 58.2 | **50.0** | 52.6 | 37.4 | 66.7 | **81.0** | **62.4** |
| | | | | *Retaining 50% Tokens* | | | | | |
| DivPrune (CVPR2025) | 45.37 | 15.7 | 61.2 | 48.6 | **53.6** | 37.7 | 65.3 | 73.0 | 62.4 |
| PACT (CVPR2025) | 45.55 | 16.9 | **61.2** | 47.2 | 48.3 | **37.9** | 62.7 | **81.0** | 61.2 |
| **EgoPrune** ($\lambda = 0.5$) | **46.16** | **16.9** | 59.7 | **50.0** | 51.7 | 36.9 | **70.7** | 73.0 | **64.7** |
| | | | | *Retaining 30% Tokens* | | | | | |
| DivPrune (CVPR2025) | 44.98 | 14.6 | **62.7** | 48.6 | **53.1** | 34.6 | 62.7 | 71.4 | 58.8 |
| PACT (CVPR2025) | 45.13 | 16.3 | 62.7 | **50.0** | 50.7 | **37.9** | 58.7 | **77.8** | 58.8 |
| **EgoPrune** ($\lambda = 0.5$) | **45.66** | **16.3** | 59.7 | 48.6 | 51.2 | 35.9 | **70.7** | 71.4 | **62.4** |



% \section{7 Appendix}

\end{document}